\documentclass{article}
\date{}
\pdfoutput=1

\newcommand{\ie}{\textit{i}.\textit{e}.}
\newcommand{\eg}{\textit{e}.\textit{g}.}
\newcommand{\etc}{\textit{etc}}

\usepackage{hyperref}       
\usepackage{url}            
\usepackage{booktabs}       
\usepackage{amsfonts}       
\usepackage{nicefrac}       
\usepackage{microtype}      
\usepackage{amssymb,amsmath,amsthm}
\usepackage{subcaption}
\usepackage{fullpage}
\usepackage{floatrow}
\usepackage{graphicx}
\usepackage{algorithm}
\usepackage{algpseudocode}

\title{Evolutionary Generative Adversarial Networks}

\author{Chaoyue Wang\textsuperscript{$\dagger$}, Chang Xu\textsuperscript{$\ddagger$}, Xin Yao\textsuperscript{$\ast\star$}, Dacheng Tao\textsuperscript{$\ddagger$}\\
\textsuperscript{$\dagger$}Centre for Artificial Intelligence, School of Software, \\
University of Technology Sydney, Australia\\
\textsuperscript{$\ddagger$}UBTECH Sydney AI Centre, School of IT, FEIT, \\
The University of Sydney, Australia
\\
\textsuperscript{$\ast$} Department of Computer Science
and Engineering, \\ Southern University of Science and Technology, China\\
\textsuperscript{$\star$} School of Computer Science, University of Birmingham, U.K.\\
chaoyue.wang@student.uts.edu.au, c.xu@sydney.edu.au\\
x.yao@cs.bham.ac.uk, dacheng.tao@sydney.edu.au}

\begin{document}
\maketitle

\begin{abstract}
Generative adversarial networks (GAN) have been effective for learning generative models for real-world data. However, existing GANs (GAN and its variants) tend to suffer from training problems such as instability and mode collapse. In this paper, we propose a novel GAN framework called evolutionary generative adversarial networks (E-GAN) for stable GAN training and improved generative performance. Unlike existing GANs, which employ a pre-defined adversarial objective function alternately training a generator and a discriminator, we utilize different adversarial training objectives as mutation operations and evolve a population of generators to adapt to the environment (i.e., the discriminator). We also utilize an evaluation mechanism to measure the quality and diversity of generated samples, such that only well-performing generator(s) are preserved and used for further training. In this way, E-GAN overcomes the limitations of an individual adversarial training objective and always preserves the best offspring, contributing to progress in and the success of GANs. Experiments on several datasets demonstrate that E-GAN achieves convincing generative performance and reduces the training problems inherent in existing GANs.
\end{abstract}

\section{Introduction}
\label{sec:introduction}
Generative adversarial networks (GAN)~\cite{goodfellow2014generative} are one of the main groups of methods used to learn generative models from complicated real-world data. As well as using a generator to synthesize semantically meaningful data from standard signal distributions, GANs (GAN and its variants) train a discriminator to distinguish \emph{real} samples in the training dataset from \emph{fake} samples synthesized by the generator. As the confronter, the generator aims to deceive the discriminator by producing ever more realistic samples. The training procedure continues until the generator wins the adversarial game; that is, the discriminator cannot make a better decision than randomly guessing whether a particular sample is fake or real. GANs have recently been successfully applied to image generation~\cite{chen2016infogan,nguyen2016plug,gan2017triangle,Zhang_2017_ICCV}, image editing~\cite{Isola_2017_CVPR,wang2017tag,zhu2017unpaired,Wang_2017_ICCV}, video prediction\cite{vondrick2016generating,finn2016unsupervised,Vondrick_2017_CVPR}, and many other tasks~\cite{zhang2016generating,oord2016wavenet,lu2017best}. 

Although GANs already produce visually appealing samples in various applications, they are often difficult to train. If the data distribution and the generated distribution do not substantially overlap (usually at the beginning of training), the generator gradients can point to more or less random directions or even result in the vanishing gradient issue. GANs also suffer from mode collapse, \ie, the generator assigns all its probability mass to a small region in the space~\cite{arora2017generalization}. In addition, appropriate hyper-parameters (\eg, learning rate and updating steps) and network architectures are critical configurations in GANs. Unsuitable settings reduce GAN's performance or even fail to produce any reasonable results. 

Many recent efforts on GANs have focused on overcoming these training difficulties by developing various adversarial training objectives. Typically, assuming the optimal discriminator for the given generator is learned, different objective functions of the generator aim to measure the distance between the data distribution and the generated distribution under different metrics. The original GAN uses Jensen-Shannon divergence as the metric. A number of metrics have been introduced to improve GAN's performance, such as least-squares~\cite{Mao_2017_ICCV}, absolute deviation~\cite{zhao2016energy}, Kullback-Leibler divergence~\cite{radford2015unsupervised,nguyen2017dual}, and Wasserstein distance~\cite{arjovsky2017wasserstein}. However, according to both theoretical analyses and experimental results, minimizing each distance has its own pros and cons. For example, although measuring Kullback-Leibler divergence largely eliminates the vanishing gradient issue, it easily results in mode collapse~\cite{radford2015unsupervised,arjovsky2017towards}. Likewise, Wasserstein distance greatly improves training stability but can have non-convergent limit cycles near equilibrium~\cite{nagarajan2017gradient}.

To exploit the advantages and suppress the weaknesses of different metrics (\ie, GAN objectives), we devise a framework that utilizes different metrics to jointly optimize the generator. In doing so, we improve both the training stability and generative performance. We build an evolutionary generative adversarial network (E-GAN), which treats the adversarial training procedure as an evolutionary problem. Specifically, a discriminator acts as the \emph{environment} (\ie, provides adaptive loss functions) and a \emph{population} of generators evolve in response to the environment. During each adversarial (or evolutionary) iteration, the discriminator is still trained to recognize real and fake samples. However, in our method, acting as parents, generators undergo different \emph{mutations} to produce offspring to adapt to the environment. Different adversarial objective functions aim to minimize different distances between the generated distribution and the data distribution, leading to different mutations. Meanwhile, given the current optimal discriminator, we measure the quality and diversity of samples generated by the updated offspring. Finally, according to the principle of ``survival of the fittest'', poorly-performing offspring are removed and the remaining well-performing offspring (\ie, generators) are preserved and used for further training.  

Based on the evolutionary paradigm to optimize GANs, the proposed E-GAN overcomes the inherent limitations in the individual adversarial training objectives and always preserves the best offspring produced by different training objectives (\ie, mutations). In this way, we contribute to progress in and the success of GANs. Experiments on several datasets demonstrate the advantages of integrating different adversarial training objectives and E-GAN's convincing performance for image generation.

\section{Related Works}
\label{sec:related}

In this section, we first review some previous GANs devoted to reducing training instability and improving the generative performance. We then briefly summarize some evolutionary algorithms on deep neural networks.

\subsection{Generative Adversarial Networks}

Generative adversarial networks (GAN) provides an excellent framework for learning deep generative models, which aim to capture probability distributions over the given data. Compared to other generative models, GAN is easily trained by alternately updating a generator and a discriminator using the back-propagation algorithm. In many generative tasks, GANs (GAN and its variants) produce better samples than other generative models~\cite{goodfellow2016nips}. 

However, some problems still exist in the GANs training process. In the original GAN, training the generator was equal to minimizing the Jensen-Shannon divergence between the data distribution and the generated distribution, which easily resulted in the vanishing gradient problem. To solve this issue, a non-saturating heuristic objective (\ie, `$-\log D$~trick') replaced the minimax objective function to penalize the generator~\cite{goodfellow2014generative}. Then, \cite{radford2015unsupervised} and \cite{salimans2016improved} designed specified network architectures (DCGAN) and proposed several heuristic tricks (\eg, feature matching, one-side label smoothing, virtual batch normalization) to improve training stability. Meanwhile, energy-based GAN~\cite{zhao2016energy} and least-squares GAN~\cite{Mao_2017_ICCV} improved training stability by employing different training objectives. Although these methods partly enhanced training stability, in practice, the network architectures and training procedure still required careful design to maintain the discriminator-generator balance. More recently, Wasserstein GAN (WGAN)~\cite{arjovsky2017wasserstein} and its variant WGAN-GP~\cite{gulrajani2017improved} were proposed to minimize the Wasserstein-1 distance between the generated and data distributions. Since the Wasserstein-1 distance is continuous everywhere and differentiable almost everywhere under only minimal assumptions~\cite{arjovsky2017wasserstein}, these two methods convincingly reduce training instability. However, to measure the Wasserstein-1 distance between the generated distribution and the data distribution, they are asked to enforce the Lipschitz constraint on the discriminator (\emph{aka} critic), which may restrict critic capability and result in some optimization difficulties~\cite{gulrajani2017improved}. 

\subsection{Evolutionary Algorithms}

Over the last twenty years, evolutionary algorithms have achieved considerable success across a wide range of computational tasks including modeling, optimization and design~\cite{eiben2003introduction,de2006evolutionary}. Inspired by natural evolution, the essence of an evolutionary algorithm is to equate possible solutions to individuals in a population, produce offspring through variations, and select appropriate solutions according to fitness~\cite{eiben2015evolutionary}.

Recently, evolutionary algorithms have been introduced to solve deep learning problems. To minimize human participation in designing deep algorithms and automatically discover such configurations, there have been many attempts to optimize deep learning hyper-parameters and design deep network architectures through an evolutionary search~\cite{young2015optimizing,miikkulainen2017evolving,real2017large}. Evolutionary algorithms have also demonstrated their capacity to optimize deep neural networks~\cite{lander2015evoae,yao1999evolving}. Moreover,~\cite{salimans2017evolution} proposed a novel evolutionary strategy as an alternative to the popular MDP-based reinforcement learning (RL) techniques, achieving strong performance on RL benchmarks. Last but not least, an evolutionary algorithm was proposed to compress deep learning models by automatically eliminating redundant convolution filters~\cite{wang2017towards}.

\begin{figure*}[!t]
\centering
\includegraphics[bb=20bp 150bp 930bp 520bp,scale=0.50]{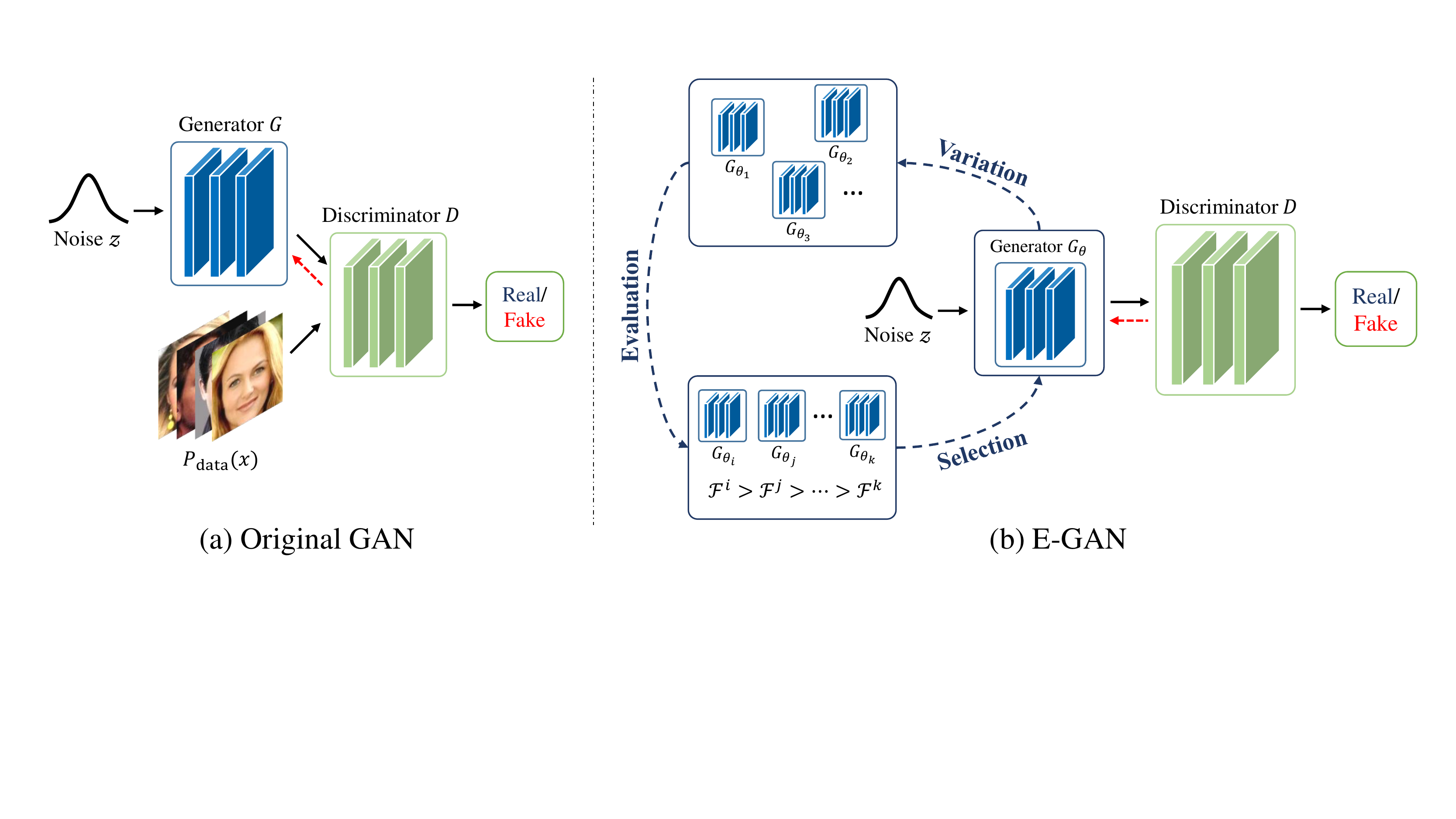}
\caption{(a) The original GAN framework. A generator $G$ and a discriminator $D$ play a two-player adversarial game. The updating gradients of the generator $G$ are received from the adaptive objective, which depends on discriminator $D$. (b) The proposed E-GAN framework. A population of generators $\{G_\theta\}$ evolves in a dynamic environment, the discriminator $D$. Each evolutionary step consists of three sub-stages: variation, evaluation, and selection. The best offspring are kept.}
\label{fig:principle}
\end{figure*}

\section{Method}
\label{sec:method}
In this section, we first review the GAN formulation. Then, we introduce the proposed E-GAN algorithm. By illustrating E-GAN's mutations and evaluation mechanism, we further discuss the advantage of the proposed framework. Finally, we conclude with the entire E-GAN training process.

\subsection{Generative Adversarial Networks}

GAN, first proposed in \cite{goodfellow2014generative}, studies a two-player minimax game between a discriminative network $D$ and a generative network $G$. Taking noisy sample $z \sim p(z)$ (sampled from a uniform or normal distribution) as the input, the generative network $G$ outputs new data $G(z)$, whose distribution $p_g$ is supposed to be close to that of the data distribution $p_\text{data}$. Meanwhile, the discriminative network $D$ is employed to distinguish the true data sample $x \sim p_\text{data}(x)$ and the generated sample $G(z) \sim p_g(G(z))$. In the original GAN, this adversarial training process was formulated as:
\begin{equation}
\min_G \max_D \mathbb{E}_{x \sim p_\text{data}}[\log D(x)] +\mathbb{E}_{z \sim p_z} [\log (1-D(G(z)))].
\end{equation}
The adversarial procedure is illustrated in Fig.~\ref{fig:principle} (a). Most existing GANs perform a similar adversarial procedure in different adversarial objective functions.

\subsection{Evolutionary Algorithm}

In contrast to conventional GANs, which alternately update a generator and a discriminator, we devise an evolutionary algorithm that evolves a population of generator(s) $\{G\}$ in a given environment (\ie, the discriminator $D$). In this population, each \emph{individual} represents a possible solution in the parameter space of the generative network $G$. During the evolutionary process, we expect that the population gradually adapts to its environment, which means that the evolved generator(s) can generate ever more realistic samples and eventually learn the real-world data distribution. As shown in Fig.~\ref{fig:principle} (b), during evolution, each step consists of three sub-stages:
\begin{itemize}
\item {\bf Variation:} Given an individual $G_\theta$ in the population, we utilize the variation operators to produce its offspring $\{G_{\theta_1}, G_{\theta_2},\cdots\}$. Specifically, several copies of each individual---or \emph{parent}---are created, each of which are modified by different \emph{mutations}. Then, each modified copy is regarded as one \emph{child}.
\item {\bf Evaluation:} For each child, its performance---or \emph{individual's quality}---is evaluated by a \emph{fitness} function $\mathcal{F}(\cdot)$ that depends on the current environment (\ie, discriminator $D$).
\item {\bf Selection:} All children will be selected according to their fitness value, and the worst part is removed---that is, they are \emph{killed}. The rest remain \emph{alive} (\ie, free to act as parents), and evolve to the next iteration.
\end{itemize}

After each evolutionary step, the discriminative network $D$ (\ie, the environment) is updated to further distinguish real samples $x$ and fake samples $y$ generated by the evolved generator(s), \ie,
\begin{equation}\label{EGAN_D}
\mathcal{L}_D = -\mathbb{E}_{x \sim p_\text{data}}[\log D(x)] - \mathbb{E}_{y \sim p_g} [\log (1-D(y))].
\end{equation}
Thus, the discriminative network $D$ (\ie, the environment) can continually provide the adaptive losses to drive the population of generator(s) evolving to produce better solutions. Next, we illustrate and discuss the proposed variation (or mutation) and evaluation operators in detail.

\subsection{Mutations}
We employ \emph{asexual reproduction} with different mutations to produce the next generation's individuals (\ie, children). Specifically, these mutation operators correspond to different training objectives, which attempt to narrow the distances between the generated distribution and the data distribution from different perspectives. In this section, we introduce the mutations used in this work\footnote{More mutation operations were tested, but the mutation approaches described already delivered a convincing performance.}. To analyze the corresponding properties of these mutations, we suppose that, for each evolutionary step, the optimal discriminator $D^*(x)=\frac{p_{data}(x)}{p_{data}(x)+p_g(x)}$, according to Eq.~(\ref{EGAN_D}), has already been learned~\cite{goodfellow2014generative}.

\subsubsection{Minimax mutation}
\begin{figure}
\begin{center}
\includegraphics[bb=50bp 225bp 530bp 615bp,scale=0.5]{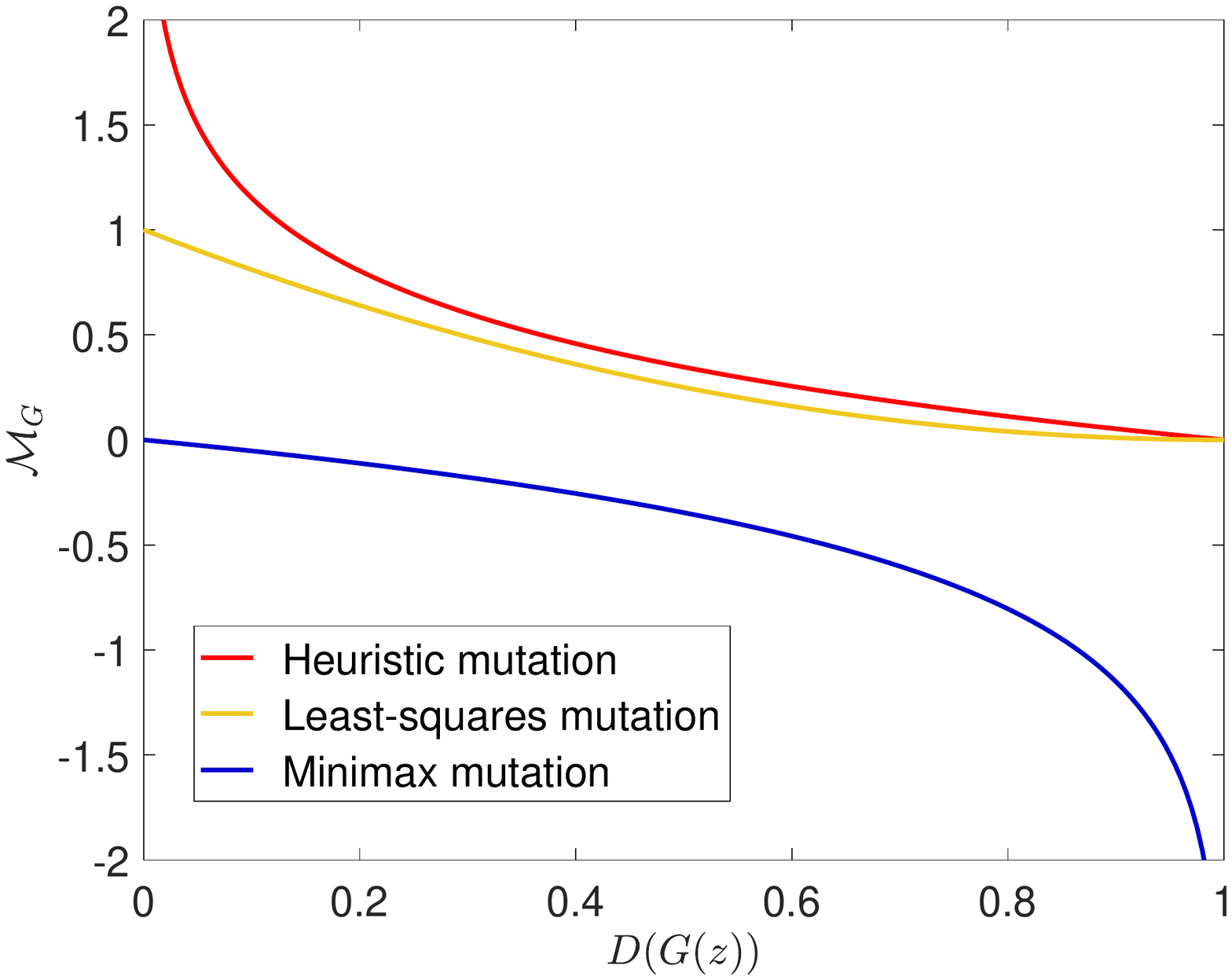}
\end{center}
  \caption{The mutation (or objective) functions that the generator $G$ receives given the discriminator $D$.}
\label{fig:lossG}
\end{figure}

The minimax mutation corresponds to the minimax objective function in the original GAN:
\begin{equation}
\mathcal{M}_G^\text{minimax} = \frac{1}{2}\mathbb{E}_{z \sim p_z} [\log (1-D(G(z))].
\end{equation}
According to the theoretical analysis in \cite{goodfellow2014generative}, given the optimal discriminator $D^*$, the minimax mutation aims to minimize the Jensen-Shannon divergence (JSD) between the data distribution and the generated distribution. Although the minimax game is easy to explain and theoretically analyze, its performance in practice is disappointing, a primary problem being the generator's vanishing gradient. If the support of two distributions lies in two manifolds, the JSD will be a constant, leading to the vanishing gradient~\cite{arjovsky2017towards}. This problem is also illustrated in Fig.~\ref{fig:lossG}. When the discriminator rejects generated samples with high confidence (\ie, $D(G(z)) \to 0$), the gradient tends to vanishing. However, if the generated distribution overlaps with the data distribution, meaning that the discriminator cannot completely distinguish real from fake samples, the minimax mutation provides effective gradients and continually narrows the gap between the data distribution and the generated distribution.

\subsubsection{Heuristic mutation}
Unlike the minimax mutation, which minimizes the log probability of the discriminator being correct, the heuristic mutation aims to maximize the log probability of the discriminator being mistaken, \ie, 
\begin{equation}
\mathcal{M}_G^\text{heuristic} = - \frac{1}{2}\mathbb{E}_{z \sim p_z} [\log (D(G(z))].
\end{equation}

Compared to the minimax mutation, the heuristic mutation will not saturate when the discriminator rejects the generated samples. Thus, the heuristic mutation avoids vanishing gradient and provides useful generator updates (Fig.~\ref{fig:lossG}). However, according to~\cite{arjovsky2017towards}, given the optimal discriminator $D^*$, minimizing the heuristic mutation is equal to minimizing $ [KL(p_g || p_\text{data}) - 2 JSD(p_g || p_\text{data})]$, \ie, inverted KL minus two JSDs. Intuitively, the JSD sign is negative, which means pushing these two distributions away from each other. In practice, this may lead to training instability and generative quality fluctuations~\cite{gulrajani2017improved}.

\subsubsection{Least-squares mutation} 
The least-squares mutation is inspired by LSGAN~\cite{Mao_2017_ICCV}, where the least-squares objectives are utilized to penalize its generator to deceive the discriminator. In this work, we formulate the least-squares mutation as:
\begin{equation}
\mathcal{M}_G^\text{least-square} = \mathbb{E}_{z \sim p_z}  [(D(G(z))-1)^2].
\end{equation}

As shown in Fig.~\ref{fig:lossG}, the least-squares mutation is non-saturating when the discriminator can recognize the generated sample (\ie, $D(G(z)) \to 0$). When the discriminator output grows, the least-squares mutation saturates, eventually approaching zero.  Therefore, similar to the heuristic mutation, the least-squares mutation can avoid vanishing gradient when the discriminator has a significant advantage over the generator. Meanwhile, compared to the heuristic mutation, although the least-squares mutation will not assign an extremely high cost to generate fake samples, it will also not assign an extremely low cost to mode dropping\footnote{\cite{arjovsky2017towards} demonstrated that the heuristic objective suffers from mode collapse since $KL(p_g || p_\text{data})$ assigns a high cost to generating fake samples but an extremely low cost to mode dropping.}, which partly avoids mode collapse~\cite{Mao_2017_ICCV}.

Note that, different from GAN-minimax and GAN-heuristic, LSGAN employs a different loss (`least-squares') from ours (Eq.~(\ref{EGAN_D})) to optimize the discriminator. Yet, as shown in the Supplementary Material, the optimal discriminator of LSGAN is equivalent to ours. Therefore, although we employ only one discriminator as the environment to distinguish real and generated samples, it is sufficient to provide adaptive losses for mutations described above.

\begin{algorithm}[t!]
  \caption{E-GAN. Default values $\alpha = 0.0002$, $\beta_1 = 0.5$, $\beta_2 = 0.99$, $n_D=2$, $n_p=1$, $n_m=3$, $m=16$.
}\label{algo:ecgan}
  \begin{algorithmic}[1]
    \Require the batch size $m$. the discriminator's updating steps per iteration $n_{D}$. the number of parents $n_p$. the number of mutations $n_m$. Adam hyper-parameters $\alpha, \beta_1, \beta_2$, the hyper-parameter $\gamma$ of evaluation function.
    \Require initial discriminator's parameters $w_0$. initial generators' parameters $\{\theta_{0}^{1}, \theta_{0}^{2}, \dots, \theta_{0}^{n_p}\}$.
    \For{number of training iterations}
      \For{$k = 0, ..., n_{D}$}

        \State Sample a batch of $\{x^{(i)}\}_{i=1}^m \sim p_\text{data}$ (training data), and a batch of $\{z^{(i)}\}_{i=1}^m \sim p_z$ (noise samples).

        \State $g_{w} \gets \nabla_{w} [\frac{1}{m}\sum_{i=1}^m \log D_w(x^{(i)})$ 

        \State \ \ \ \ $+ \frac{1}{m}\sum_{j=1}^{n_p} \sum_{i=1}^{m/n_p} \log(1 - D_w(G_{\theta^{j}}(z^{(i)}))) ]$

        \State $w \gets \text{Adam}(g_w, w, \alpha, \beta_1, \beta_2)$
      \EndFor

      \For{$j = 0, ..., n_p$}
            \For{$h = 0, ..., n_m$}
                  \State Sample a batch of $\{z^{(i)}\}_{i=1}^m \sim p_z$
                  \State $g_{\theta^{j,h}} \gets \nabla_{\theta^j} \mathcal{M}_G^h(\{z^{(i)}\}_{i=1}^m, \theta^j)$
                  \State $\theta_\text{child}^{j,h}\gets \text{Adam}(g_{\theta^{j,h}}, \theta^j, \alpha, \beta_1, \beta_2)$
                  \State $\mathcal{F}^{j,h} \gets \mathcal{F}_q^{j,h} + \gamma\mathcal{F}_d^{j,h}$
            \EndFor
      \EndFor
\State $ \{\mathcal{F}^{j_1,h_1},\mathcal{F}^{j_2,h_2},\dots\} \gets \text{sort}(\{\mathcal{F}^{j,h}\})$
\State $\theta^1,\theta^2, \dots, \theta^{n_p} \gets \theta_\text{child}^{j_1,h_1}, \theta_\text{child}^{j_2,h_2}, \dots, \theta_\text{child}^{j_{n_p},h_{n_p}}$
\EndFor
\end{algorithmic}
\end{algorithm}

\subsection{Evaluation}

In an evolutionary algorithm, evaluation is the operation of measuring the quality of individuals.  To determine the evolutionary direction (\ie, individuals' selection), we devise an evaluation (or fitness) function to measure the performance of evolved individuals (\ie, children). Typically, we focus on two generator properties:  1) the quality and 2) the diversity of generated samples. First, we simply feed generator produced images into the discriminator $D$ and observe the average value of the output, which we name the \emph{quality fitness score}:
\begin{equation}
\mathcal{F}_\text{q} = \mathbb{E}_{z} [D(G(z))].
\end{equation} 
Note that discriminator $D$ is constantly upgraded to be optimal during the training process, reflecting the quality of generators at each evolutionary (or adversarial) step. If a generator obtains a relatively high quality score, its generated samples can deceive the discriminator and the generated distribution is further approximate to the data distribution.

Besides generative quality, we also pay attention to the diversity of generated samples and attempt to overcome the mode collapse issue in GAN optimization. Recently,~\cite{nagarajan2017gradient} proposed a gradient-based regularization term to stabilize the GAN optimization and suppress mode collapse. Through their observation, when the generator collapses to a small region, the discriminator will subsequently label collapsed points as fake with obvious countermeasure (\ie, big gradients).

We employ a similar principle to evaluate generator optimization stability and generative diversity. Formally, the \emph{diversity fitness score} is defined as:
\begin{equation}
\mathcal{F}_\text{d} = -\log \big|\big|\nabla_{D} -\mathbb{E}_{x}[\log D(x)] - \mathbb{E}_{z} [\log (1-D(G(z)))]\big|\big|.
\end{equation}

The log gradient value of updating $D$ is utilized to measure the diversity of generated samples. If the updated generator obtains a relatively high diversity score, which corresponds to small discriminator gradients, its generated samples tend to spread out enough, to avoid the discriminator has obvious countermeasures. Thus, the mode collapse issue can be suppressed and the discriminator will change smoothly, which helps to improve the training stability.

\begin{figure}
\begin{center}
\includegraphics[bb=260bp 75bp 600bp 420bp,scale=0.485]{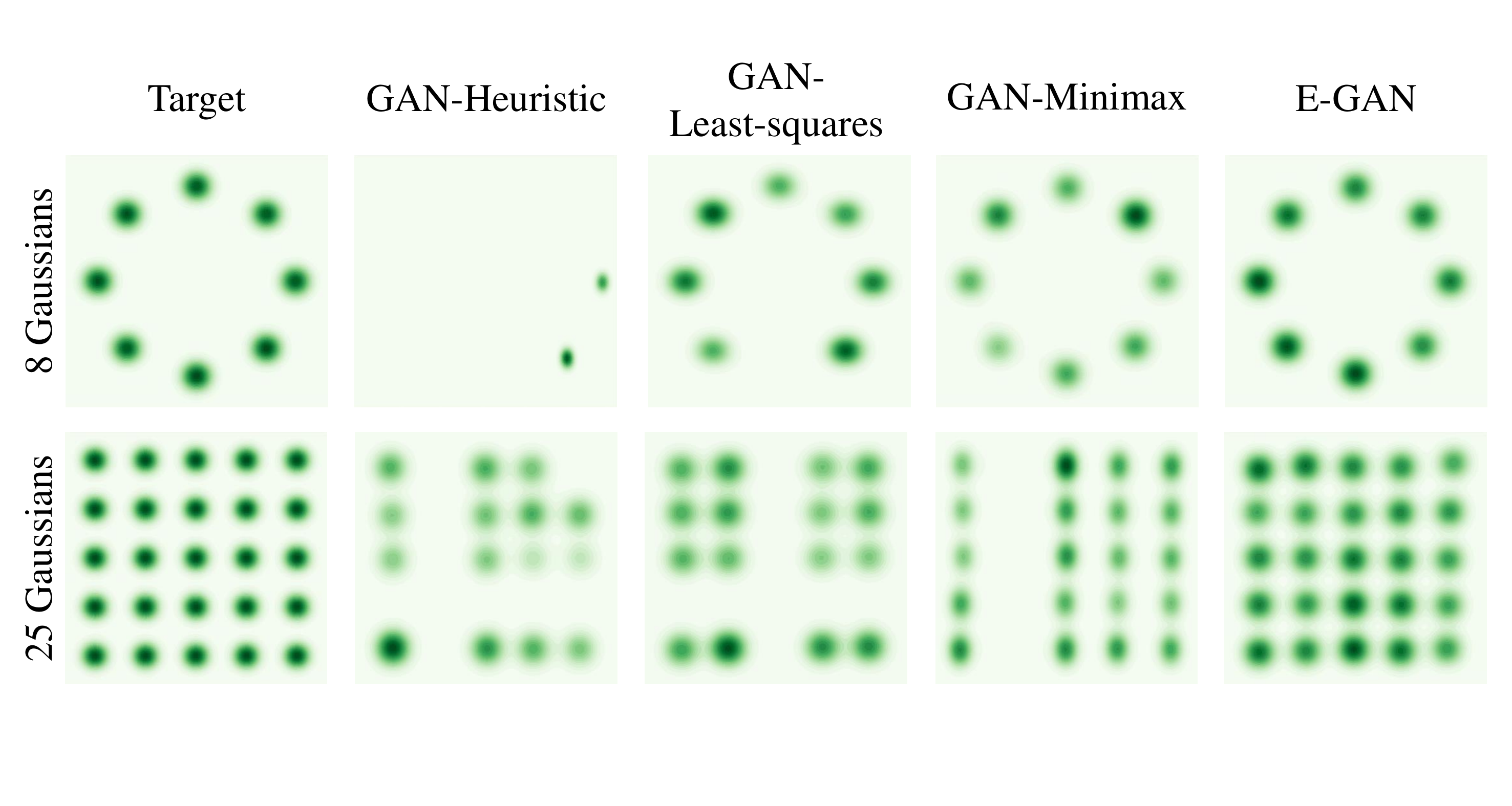}
\end{center}
  \caption{KDE plots of the target data and generated data from different GANs trained on mixtures of Gaussians. }
\label{fig:toy}
\end{figure}

\begin{figure*}[!t]
\captionsetup[subfigure]{labelformat=empty}
\centering
 \begin{subfigure}[b]{0.32\textwidth}
    \includegraphics[bb=45bp 220bp 595bp 470bp,scale=0.265]{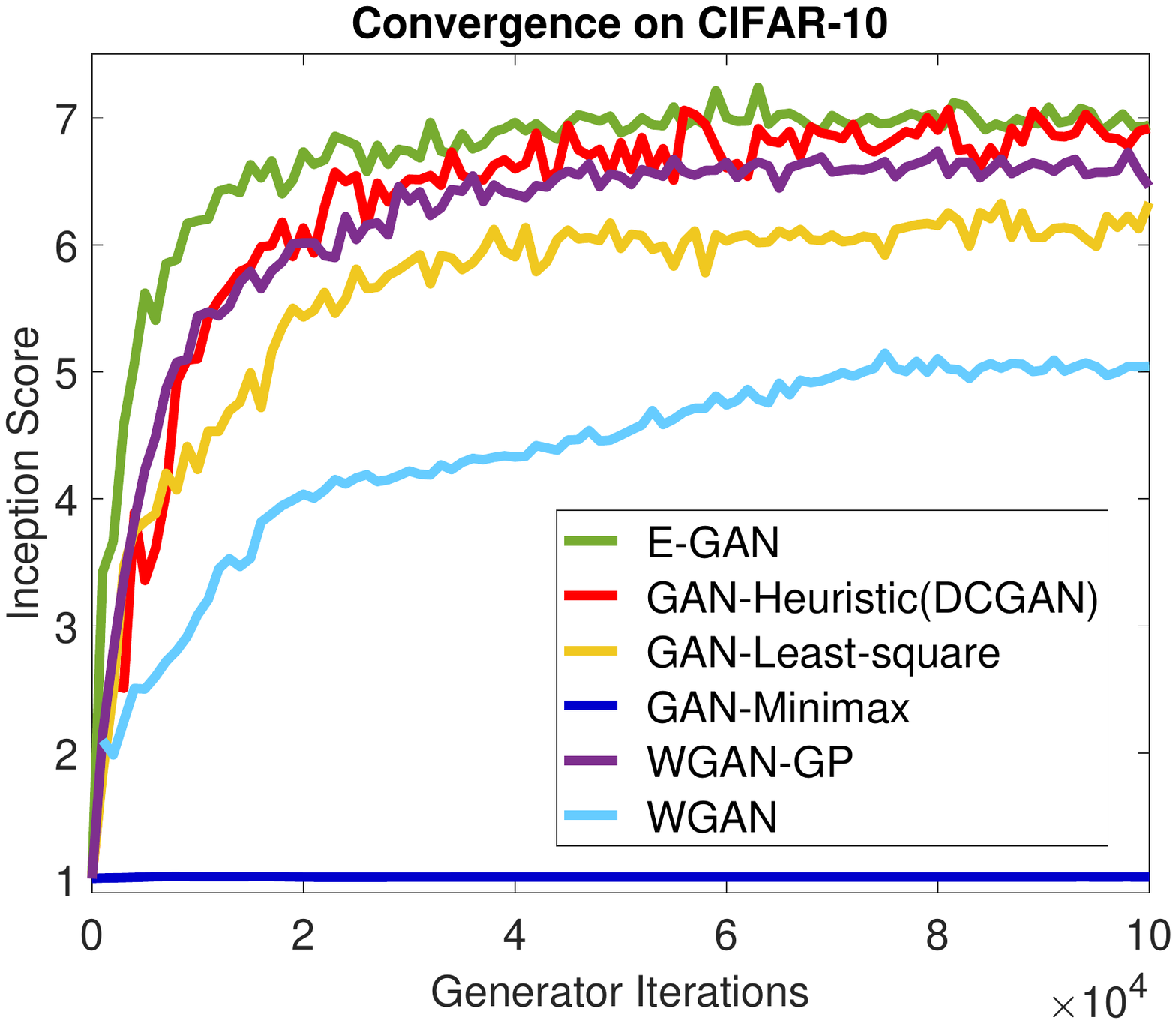}%
    \label{fig_first_case}
 \end{subfigure}
 \begin{subfigure}[b]{0.32\textwidth}
    \includegraphics[bb=44bp 220bp 595bp 470bp,scale=0.265]{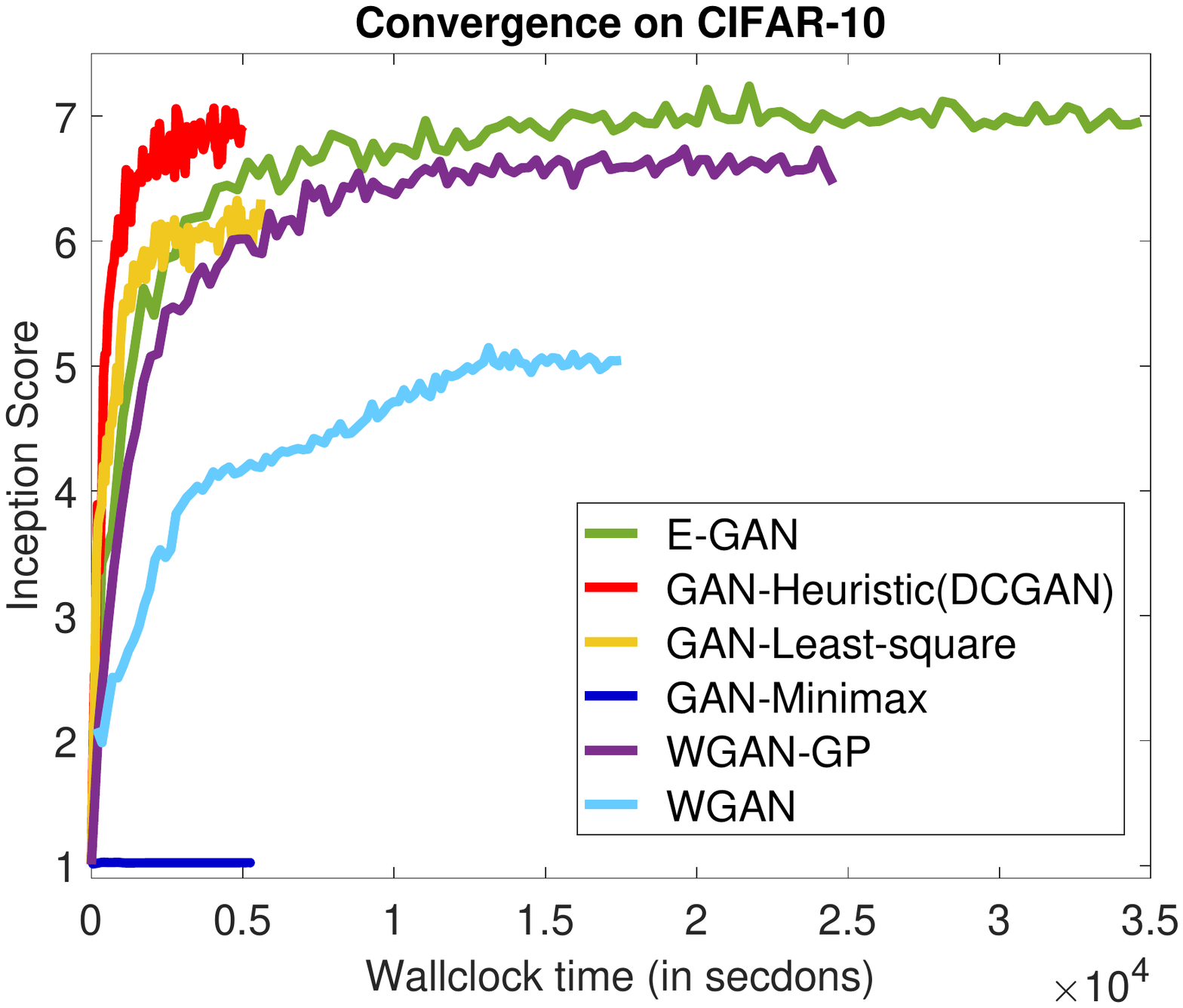}%
    \label{fig_first_case}
 \end{subfigure}
 \begin{subfigure}[b]{0.33\textwidth}
    \includegraphics[bb=65bp 95bp 1000bp 470bp,scale=0.29]{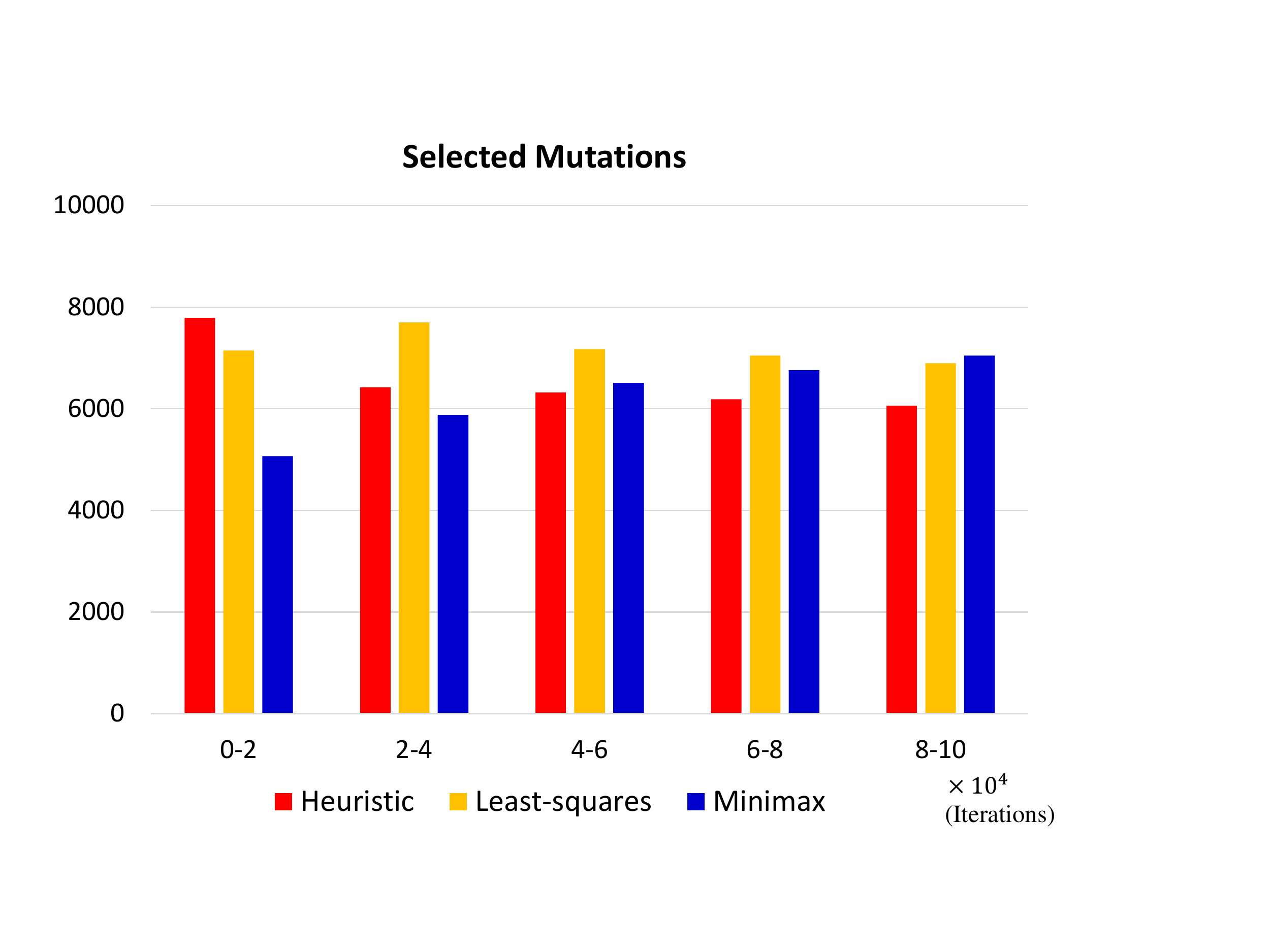}%
    \label{fig_first_case}
 \end{subfigure}
\caption{Experiments on the CIFAR-10 dataset. CIFAR-10 inception score over generator iterations (left), over wall-clock time (middle), and the graph of selected mutations in the E-GAN training process (right).}
\label{fig:analysis}
\end{figure*}

Based on the aforementioned two fitness scores, we can finally give the evaluation (or fitness) function of the proposed evolutionary algorithm:
\begin{equation}
\mathcal{F} =\mathcal{F}_\text{q}+\gamma\mathcal{F}_\text{d},
\end{equation}
where $\gamma \geq 0$ balances two measurements: generative quality and diversity. Overall, a relatively high fitness score $\mathcal{F}$, leads to higher training efficiency and better generative performance.

\subsection{E-GAN}

Having introduced the proposed evolutionary algorithm and corresponding mutations and evaluation criteria, the complete E-GAN training process is concluded in Algorithm~\ref{algo:ecgan}. Overall, in E-GAN, generators $\{G\}$ are regarded as an evolutionary population and discriminator $D$ acts as an environment. For each evolutionary step, generators are updated with different objectives (or mutations) to accommodate the current environment. According to the principle of ``survival of the fittest'', only well-performing children will survive and participate in future adversarial training. Unlike the two-player game with a fixed and static adversarial training objective in conventional GANs, E-GAN allows the algorithm to integrate the merits of different adversarial objectives and generate the most competitive solution. Thus, during training, the evolutionary algorithm not only largely suppresses the limitations (vanishing gradient, mode collapse, \etc.) of individual adversarial objectives, but it also harnesses their advantages to search for a better solution.

\section{Experiments}

To evaluate the proposed E-GAN, in this section, we run and analyze experiments on several generation tasks.

\subsection{Implementation Details}
 
We evaluate E-GAN on two synthetic datasets and three image datasets: CIFAR-10~\cite{krizhevsky2009learning}, LSUN bedroom~\cite{yu2015lsun}, and CelebA~\cite{liu2015deep}. For all of these tasks, the network architectures are based on DCGAN~\cite{radford2015unsupervised} and are briefly introduced here, more details can be found in the Supplementary Material. We use the default hyper-parameter values listed in Algorithm~\ref{algo:ecgan} for all experiments. Note that the number of parents $n_p$ is set as 1, which means only one (\ie, the best) child is retained in each evolutionary step. On the one hand, this reduces E-GAN's computational cost, thereby accelerating training. On the other, our experiments show that E-GAN already achieves impressive performance and stability even with only one survivor at each step. Furthermore, all experiments were trained on Nvidia GTX 1080Ti GPUs. To train a model for $64\times64$ images using the DCGAN architecture cost around 30 hours on a single GPU.

\subsection{Synthetic Datasets and Mode Collapse}

In the first experiment, we adopt the experimental design proposed in~\cite{metz2016unrolled}, which trains GANs on 2D Gaussian mixture distributions. The mode collapse issue can be accurately measured on these synthetic datasets, since we can clearly observe the data distribution and the generated distribution. As shown in Fig.~\ref{fig:toy}, we employ two challenging distributions to evaluate E-GAN, a mixture of 8 Gaussians arranged in a circle and a mixture of 25 Gaussians arranged in a grid.\footnote{We obtain both 2D distributions and network architectures from the code provided in~\cite{gulrajani2017improved}.}

We first compare the proposed evolutionary adversarial training framework with one using an individual adversarial objective (\ie, conventional GANs). We train each method 50K iterations and report the KDE plots in Fig.~\ref{fig:toy}. The results show that all of the individual adversarial objectives suffer from mode collapse to a greater or lesser degree. However, by combining different objectives in our evolution framework, model performance is largely improved and can accurately fit the target distributions. This further demonstrates, during the evolutionary procedure, the proposed evaluation mechanism can recognize well-performing updatings (\ie, offspring), and promote the population to a better evolutionary direction.

\begin{figure}
\begin{center}
\includegraphics[width=0.6\textwidth]{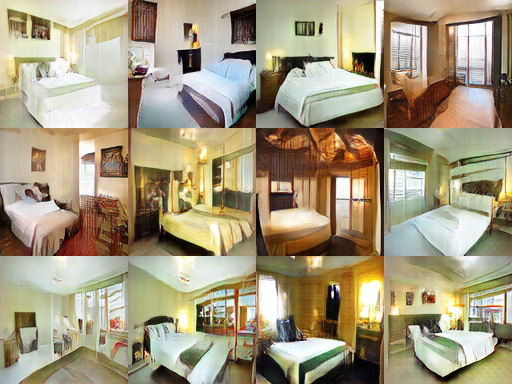}
\end{center}
  \caption{Generated samples on $128\times128$ LSUN bedrooms.}
\label{fig:beds}
\end{figure}

\subsection{CIFAR-10 and Inception Score}

When evaluating a GAN model, sample quality and convergence speed are two important criteria. We train different GANs on CIFAR-10 and plot inception scores~\cite{salimans2016improved} over the course of training (Fig.~\ref{fig:analysis}-left, middle). The same network architecture based on DCGAN is used in all methods. 

As shown in Fig.~\ref{fig:analysis}-left, E-GAN can get higher inception score with less training steps. Meanwhile, E-GAN also shows comparable stability when it goes to convergence. By comparison, conventional GANs expose their different limitations, such as instability at convergence (GAN-Heuristic), slow convergence (GAN-Least square) and invalid (GAN-minimax). As mentioned above, different objectives aim to measure the distance between the generated and data distributions under different metrics which have different pros and cons. Here, utilizing the evolutionary framework, E-GAN not only overcomes the limitations of these individual adversarial objectives, but it also outperforms other GANs (the WGAN and its improved variation WGAN-GP). Furthermore, although E-GAN takes more time for each iteration, it achieves comparable convergence speed in terms of wall-clock time (Fig.~\ref{fig:analysis}-middle).

During training E-GAN, we recorded the selected objective in each evolutionary step (Fig.~\ref{fig:analysis}-right). At the beginning of training, the heuristic mutation and the least-square mutation are selected more frequently than the minimax mutation. It may due to the fact that the minimax mutation is hard to provide effective gradients (\ie, vanishing gradient) when the discriminator can easily recognize generated samples. Along with the generator approaching convergence (after 20K steps), ever more minimax mutations are employed, yet the number of selected heuristic mutations is falling. As aforementioned, the minus JSDs of the heuristic mutation may tend to push the generated distribution away from data distribution and lead to training instability. However, in E-GAN, beyond the heuristic mutation, we have other options of mutation, which improves the stability at convergence.

\subsection{LSUN and Architecture Robustness}

\begin{figure*}[!t]
\begin{center}
\includegraphics[bb=4bp 208bp 900bp 710bp,scale=0.59]{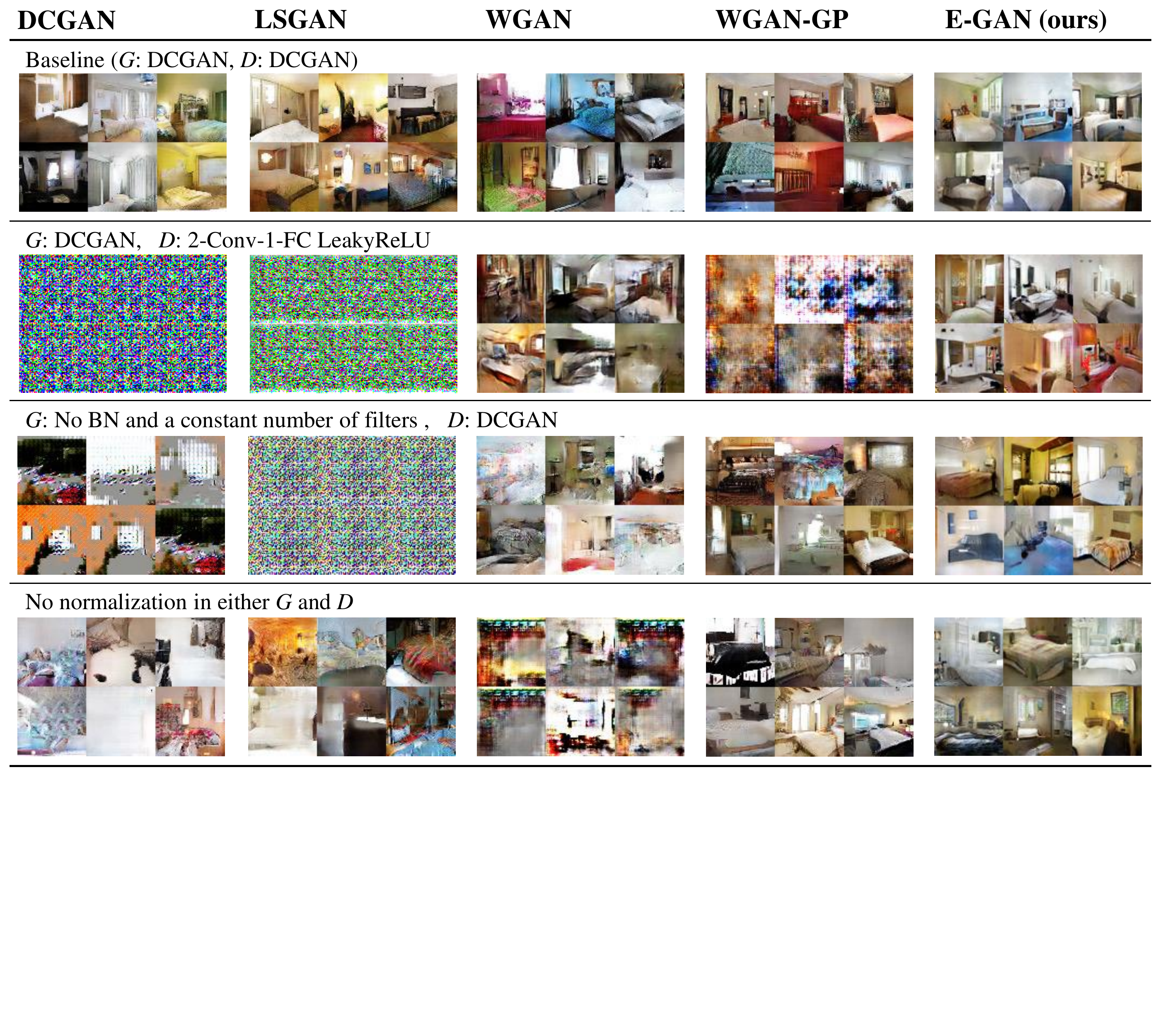}
\end{center}
  \caption{Experiments to test architecture robustness. Different GAN architectures corresponding to different training challenges and trained with five different GAN methods.}
\label{fig:bedrooms}
\end{figure*}

The architecture robustness is another advantage of E-GAN. To demonstrate the training stability of our method, we train different network architectures on the LSUN bedroom dataset~\cite{yu2015lsun} and compare with several existing works. In addition to the baseline DCGAN architecture, we choose three additional architectures corresponding to different training challenges: (1) limiting the recognition capability of the discriminator $D$, \ie, 2-Conv-1-FC LeakyReLU discriminator; (2) limiting the expression capability of the generator $G$, \ie, no batchnorm and a constant number of filters in the generator; (3)  reducing the network capability of the generator and discriminator together, \ie, remove the BN in both the generator $G$ and discriminator $D$. For each architecture, we test five different methods: DCGAN, LSGAN, standard WGAN (with weight clipping), WGAN-GP (with gradient penalty) ,and our E-GAN. For each method, we used the default configurations recommended in the respective studies (these methods are summarized in~\cite{gulrajani2017improved}) and train each model for 200K iterations. As shown in Fig.~\ref{fig:bedrooms}, E-GAN generates reasonable results even when other methods are failed. Furthermore, based on the DCGAN architecture, we train E-GAN to generate $128\times128$ bedroom images\footnote{We remove batchnorm layers in the generator. The detailed architecture and more generated images are reported in the Supplementary Material.} (Fig.~\ref{fig:beds}). Observing generated images, we demonstrate that E-GAN can be trained to generate diversity and high-quality images from the target data distribution. 

\subsection{CelebA and Space Continuity}

\begin{figure}
\begin{center}
\includegraphics[width=0.7\textwidth]{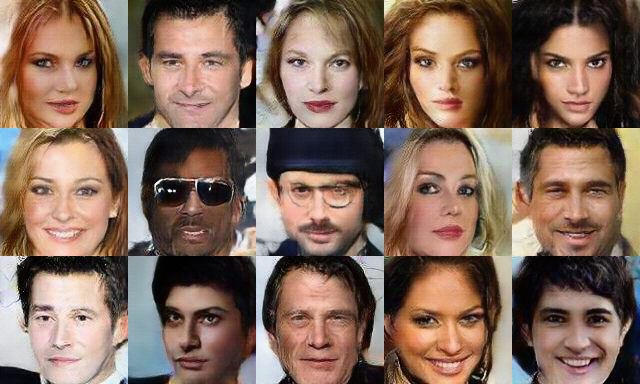}
\end{center}
  \caption{Generated human face images on the $128\times128$ CelebA dataset.}
\label{fig:face}
\end{figure}

\begin{figure*}[!t]
\centering
\includegraphics[bb=8bp 37bp 1090bp 557bp,scale=0.44]{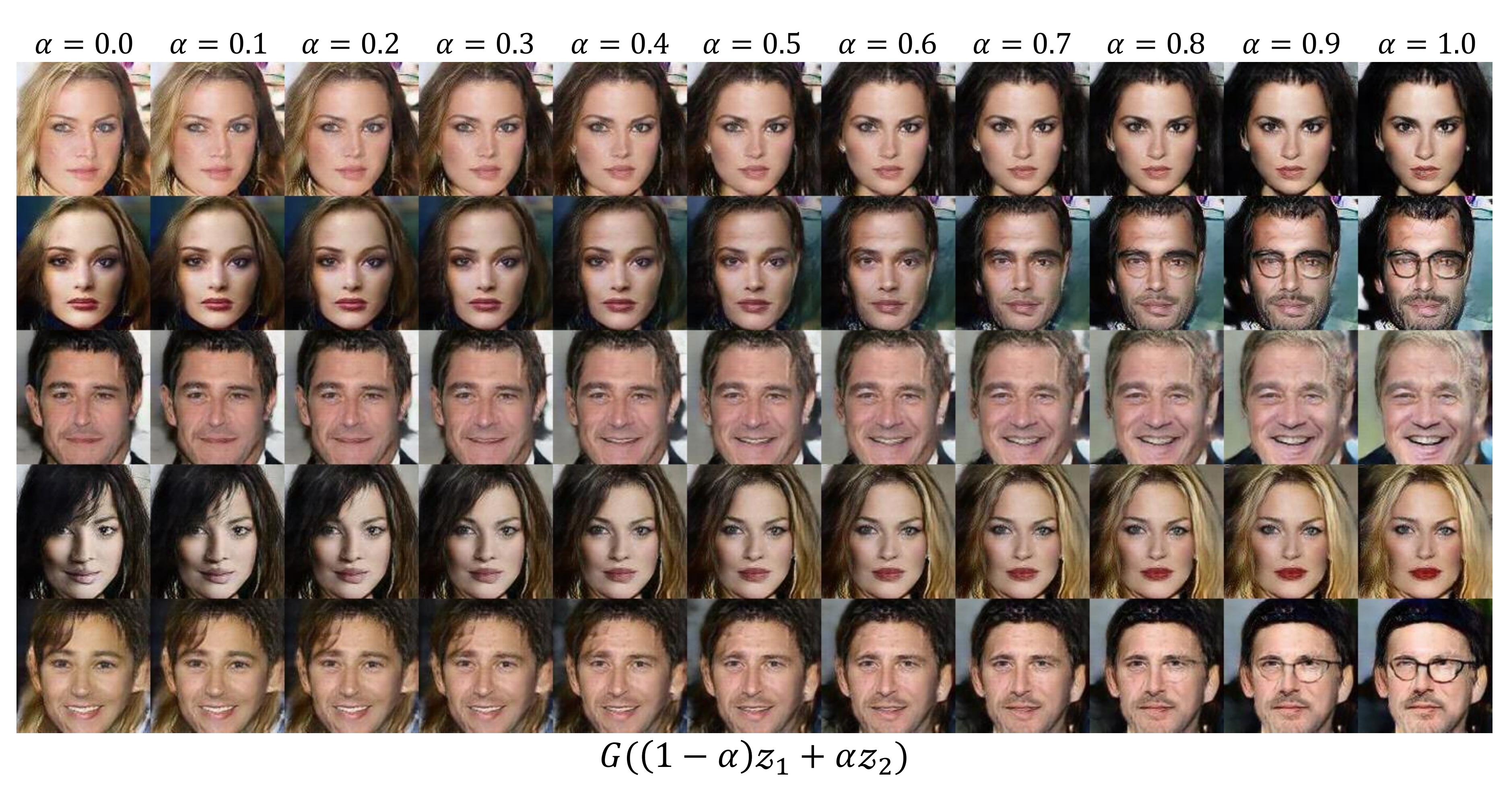}
\caption{Interpolating in latent space. For selected pairs of generated images from a well-trained E-GAN model, we record their latent vectors $z_1$ and $z_2$. Then, samples between them are generated by linear interpolation between these two vectors.}
\label{fig:face_latent}
\end{figure*}

Since humans excel at identifying facial flaws, generating high-quality human face images is challenging. Similar to generating bedrooms, we employ the same architectures to generate $128\times128$ RGB human face images (Fig.~\ref{fig:face}). In addition, given a well-trained generator, we evaluate the performance of the embedding in the latent space of noisy vectors $z$. In Fig.~\ref{fig:face_latent}, we first select pairs of generated faces and record their corresponding latent vectors $z_1$ and $z_2$. The two images in one pair have different attributes, such as gender, expression, hairstyle, and age. Then, we generate novel samples by linear interpolating between these pairs (\ie, corresponding noisy vectors). We find that these generated samples can seamlessly change between these semantically meaningful face attributes. This experiment demonstrates that generator training does not merely memorize training samples but learns a meaningful projection from latent noisy space to face images. Meanwhile, it also shows that the generator trained by E-GAN does not suffer from mode collapse, and shows great space continuity. 

\section{Conclusion}

In this paper, we present an evolutionary GAN framework (E-GAN) for training deep generative models. To reduce training difficulties and improve generative performance, we devise an evolutionary algorithm to evolve a population of generators to adapt to the dynamic environment (\ie, the discriminator $D$). In contrast to conventional GANs, the evolutionary paradigm allows the proposed E-GAN to overcome the limitations of individual adversarial objectives and preserve the best offspring after each iteration. Experiments show that E-GAN improves the training stability of GAN models and achieves convincing performance in several image generation tasks. Future works will focus on further exploring the relationship between the environment (\ie, discriminator) and evolutionary population (\ie, generators) and further improving generative performance.

\bibliographystyle{abbrv}
\bibliography{main}

\end{document}